\documentclass{article}

\usepackage[preprint]{neurips_2025}
\usepackage{hyperref} % 引入 hyperref 包来使用 \href 命令

\usepackage[utf8]{inputenc} % allow utf-8 input
\usepackage[T1]{fontenc}    % use 8-bit T1 fonts
\usepackage{newunicodechar}
\newunicodechar{─}{\textemdash{}}
\usepackage{hyperref}       % hyperlinks
\usepackage{url}            % simple URL typesetting
\usepackage{booktabs}       % professional-quality tables
\usepackage{amsfonts}       % blackboard math symbols
\usepackage{comment}       % enables the comment environment
\usepackage{nicefrac}       % compact symbols for 1/2, etc.
\usepackage{microtype}      % microtypography
\usepackage{xcolor}         % colors
\usepackage{amsmath}
\usepackage{physics}
\usepackage{bm}
\usepackage{multirow}
\usepackage{amssymb}
\usepackage{graphicx}
\usepackage{subcaption}

\usepackage{colortbl}
\usepackage{color}

\usepackage{amsthm}

\newtheorem{theorem}{Theorem}
\newtheorem{corollary}[theorem]{Corollary}

\definecolor{mygray}{gray}{.9}
\definecolor{light-gray}{gray}{0.5}
\definecolor{pretty-blue}{RGB}{0, 113, 188}
\definecolor{linecolor1}{gray}{.95} % soft gray
\definecolor{linecolor}{gray}{.895} % soft gray
\definecolor{red}{rgb}{1,0,0} % soft gray

\usepackage[capitalize]{cleveref}
\Crefname{section}{Section}{Sections}
\Crefname{table}{Table}{Tables}
\crefname{figure}{Figure}{Figures}
\crefname{equation}{Equation}{Equations}

\PassOptionsToPackage{sort&compress}{natbib}

\title{Dynamic Manipulation of Deformable Objects in 3D: Simulation, Benchmark and Learning Strategy}
\author{
  \begin{tabular}{c}
    Guanzhou Lan\textsuperscript{1}\footnotemark[1] \quad Yuqi Yang\textsuperscript{1}\footnotemark[1] \quad 
    Anup T. Mathew\textsuperscript{2} \quad Feiping Nie\textsuperscript{1} \quad Rong Wang\textsuperscript{1} \\ \ Xuelong Li\textsuperscript{3} \quad 
    Federico Renda\textsuperscript{2}\footnotemark[2] \quad Bin Zhao\textsuperscript{1}\footnotemark[2] \\
    \textsuperscript{1}Northwestern Polytechnical University \quad
   \textsuperscript{2}Khalifa University \quad
   \textsuperscript{3}TeleAI
  \end{tabular}
}

\begin{document}

\maketitle
{
\renewcommand{\thefootnote}{\fnsymbol{footnote}}
\footnotetext[1]{Equal Contribution.}
}

{
\renewcommand{\thefootnote}{\fnsymbol{footnote}}
\footnotetext[2]{Corresponding author.}
}

\begin{abstract}
Goal-conditioned dynamic manipulation is inherently challenging due to complex system dynamics and stringent task constraints, particularly in deformable object scenarios characterized by high degrees of freedom and underactuation. Prior methods often simplify the problem to low-speed or 2D settings, limiting their applicability to real-world 3D tasks. In this work, we explore 3D goal-conditioned rope manipulation as a representative challenge. To mitigate data scarcity, we introduce a novel simulation framework and benchmark grounded in reduced-order dynamics, which enables compact state representation and facilitates efficient policy learning. Building on this, we propose \textbf{D}ynamics \textbf{I}nformed \textbf{D}iffusion \textbf{P}olicy (\textbf{DIDP}), a framework that integrates imitation pretraining with physics-informed test-time adaptation. First, we design a diffusion policy that learns inverse dynamics within the reduced-order space, enabling imitation learning to move beyond naïve data fitting and capture the underlying physical structure. Second, we propose a physics-informed test-time adaptation scheme that imposes kinematic boundary conditions and structured dynamics priors on the diffusion process, ensuring consistency and reliability in manipulation execution. Extensive experiments validate the proposed approach, demonstrating strong performance in terms of accuracy and robustness in the learned policy.
\end{abstract}

\section{Introduction}

Goal-conditioned dynamic manipulation poses significant challenges in robotics, particularly when dealing with deformable objects. The first challenge stems from the inherently high-dimensional dynamics of such objects, which involve numerous degrees of freedom (DoF), making accurate modeling and real-time control computationally demanding. The second challenge stems from the requirement for high-precision control in underactuated systems: minor deviations in force, contact conditions, or trajectory can result in task failure due to the inherently sensitive and nonlinear behavior of soft materials.

% 这里要重新写一下（1）过去的方法都是FEM的仿真，纯数值方法，参数量很大，所以在2D任务，xxx xxx xxx xxx xxx，可以减小维度的参数量，2维空间的learning也可以借鉴CV里的learning方法，不太困难。（2）也有工作尝试了3维的探索，但是制作了仿真的探索。

% 一旦问题上升到三维，(1)维度的提升带来巨大的参数量。（2）三维的数据结构导致了单纯data fitting 学到make sense的表征更加困难。因此，需要从建模，仿真，到学习方法，进行comprehensive exploration/
Previous efforts in deformable object manipulation primarily fall into two categories. The first line of work relies on numerical simulations based on Finite Element Methods (FEM), which involve large numbers of parameters and are computationally intensive. To mitigate this complexity, many studies restrict the manipulation to 2D settings, where the reduced dimensionality allows for more tractable modeling and enables the use of learning techniques from artificial intelligence \cite{li2015folding, jangir2020dynamic,hietala2022learning}. For example, iterative learning frameworks have been proposed to augment the input space while reducing the underlying parameter dimensionality~\cite{chi2024iterative}, enabling effective policy learning from $256 \times 256$ image sequences. However, these methods are inherently limited to planar tasks.
Few research has attempted to explore dynamic behaviors in 3D~\cite{nah2021manipulating}, but often relies on high-dimensional simulation models, such as 50-DoFs representations, to describe system dynamics, which introduces substantial computational overhead and struggles to learn physically consistent dynamics in high-dimensional spaces.

Compared to 2D tasks, 3D goal-conditioned dynamic manipulation of deformable objects poses significantly greater challenges due to: (1) increased dimensionality and parameter complexity required to accurately model the system; and (2) the difficulty of learning meaningful and generalizable policies from sparse and high-dimensional data. These challenges call for a comprehensive framework that integrates efficient modeling, scalable simulation, and physically consistent learning approaches.

To address these challenges, we first adopt the reduced-order Geometric Variable Strain (GVS) model~\cite{Anup2025reduced} to construct the simulation environment and associated benchmarks. This model (1) significantly reduces the dimensionality of deformable object representation, requiring only 20 DoF, thereby decreasing the number of parameters by over 50\% for efficient modeling and simulation; and (2) offers a unified and differentiable dynamics formulation that jointly models the kinematics and dynamics of rigid manipulators and deformable objects within the same framework, thereby facilitating end-to-end optimization and physically consistent policy learning.
Building on this foundation, we propose the \textbf{D}ynamics-\textbf{I}nformed \textbf{D}iffusion \textbf{P}olicy (\textbf{DIDP}), a novel framework that integrates diffusion-based pretraining with physics-informed test-time adaptation. Unlike existing approaches, our diffusion policy learns inverse dynamics within the reduced-order space, enabling the model to capture full-system inverse dynamics beyond mere data fitting. Furthermore, a physics-informed test-time adaptation mechanism is introduced, which incorporates kinematic boundary constraints and differentiable dynamics priors into the diffusion process, thereby ensuring physically consistent and reliable manipulation outcomes. In summary, our contributions are as follows:

(1) We introduce a reduced-order GVS model for 3D dynamic manipulation of deformable objects, enabling the construction of the first simulation environment and benchmark with only 20 DoF. % This model provides an efficient yet expressive representation of soft body dynamics.

(2) We develop the DIDP framework to achieve 3D goal-conditioned dynamic manipulation. Unlike conventional behavior cloning approaches, our method explicitly models the full inverse dynamics, enhancing both interpretability and generalization

(3) We derive a differentiable dynamics prior from the GVS model to enable an end-to-end test-time adaptation strategy, facilitating physically consistent policy learning.

\begin{figure*}[t]
\centering
\includegraphics[width=0.95\linewidth]{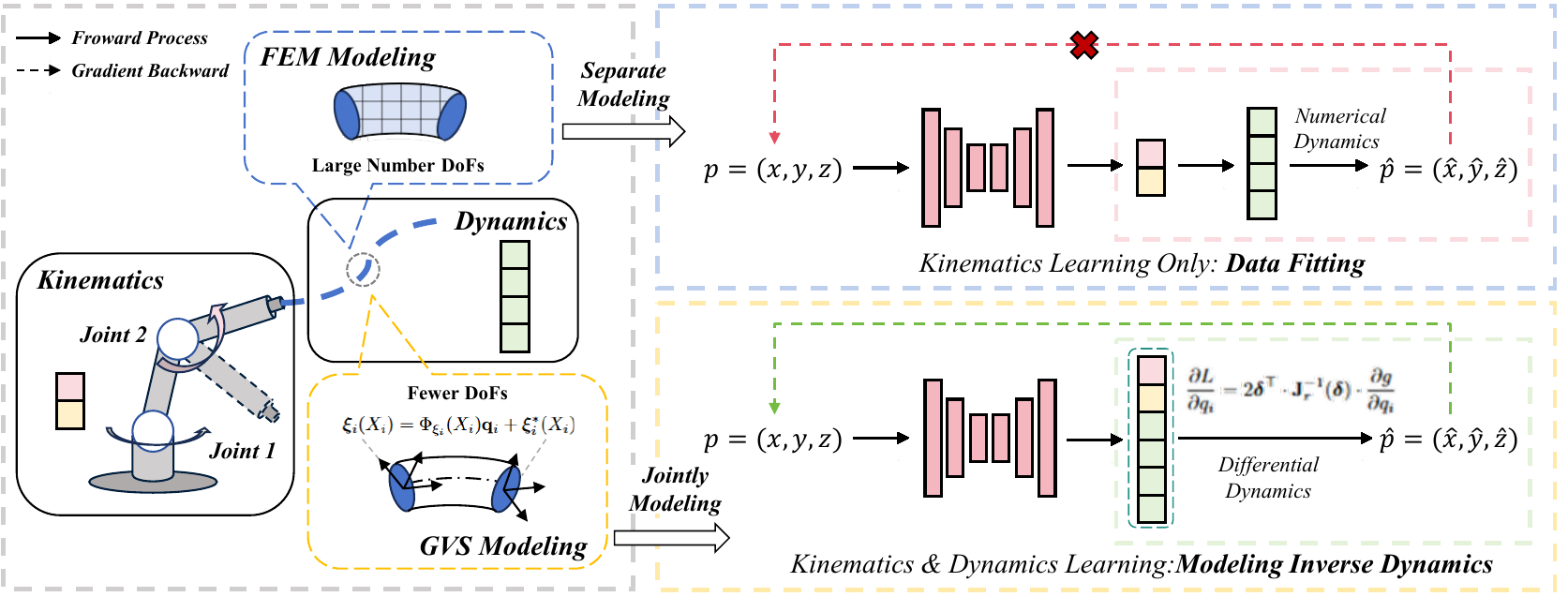} 
\vspace{-0.3cm}
\caption{We introduce the 3D dynamic manipulation benchmark for deformable objects by leveraging GVS modeling. This framework enables a reduced-DoF representation and differentiable system modeling, facilitating efficient learning of inverse dynamics. Our approach ensures consistency between kinematics and dynamics throughout the policy learning process.}
\vspace{-0.5cm}
\label{fig:top1}
\end{figure*}

\section{Related Work}
% 经典方法困境：揭示物理建模与优化的根本性计算瓶颈
% 数据驱动局限：指出学习范式因忽视物理先验导致的低效问题
% 新兴技术缺口：剖析扩散模型在动态物理约束建模中的不足
% → 最终锚定创新点：物理引导的降维扩散框架，同步突破三个维度的限制。

\subsection{Deformable Object Manipulation}
Early work in deformable object manipulation focused on quasi-static scenarios, often limited to 2D planes. These methods rely on vision-based inputs and hand-engineered dynamics tailored to specific tasks, such as cloth folding or rope knotting \cite{seita2020deep, hoque2022visuospatial, yin2021modeling}. While effective in constrained settings, they fall short in generalizing to high-speed or 3D environments.

Data-driven approaches have introduced learned representations of deformable states, often from visual observations or multimodal sensing (e.g., haptics, audio) \cite{grannen2021untangling, huang2023self, li2023see}. These enable tasks like cable reshaping and cloth folding \cite{jin2022robotic, xu2022dextairity}, but typically assume full observability or rely on expert demonstrations, limiting adaptability in dynamic or partially observed environments.

Unlike quasi-static tasks, dynamic manipulation leverages inertia and transient dynamics for high-speed motions such as whipping, throwing, or catching \cite{mason1986mechanics, jangir2020dynamic}. Optimization-based methods (e.g., iLQR, CHOMP) offer precision but are sensitive to modeling errors and lack robustness in real-world settings \cite{li2015folding, toussaint2018differentiable}. Iterative Learning Control (ILC) methods, like IRP \cite{chi2024iterative}, improve generalization via residual feedback but assume task repeatability and known Jacobians.

\subsection{Diffusion-Based Robotic Manipulation}
Diffusion models \cite{ho2020denoising} have emerged as a powerful framework in generative modeling, capable of accurately capturing complex, high-dimensional data distributions. They have achieved notable success in diverse domains such as image generation \cite{rombach2022high,peebles2023scalable}, video generation \cite{wang2025lavie,ma2024latte}, 3D shape synthesis \cite{pooledreamfusion}, and robotic policy learning \cite{chi2023diffusion,ze20243d}. 

In the context of robotic planning, particularly within complex and high-dimensional environments, diffusion-based generative models have been adopted to produce goal-conditioned action trajectories that better capture the stochasticity and flexibility needed for dynamic control \cite{chi2023diffusion, luo2024text}. Hierarchical approaches, such as the Hierarchical Diffusion Policy (HDP), further enhance this capability by coupling high-level task decomposition with kinematically-aware, low-level diffusion control, achieving competitive performance across a variety of manipulation tasks \cite{Ma_2024_CVPR}.

However, existing applications of diffusion policies have predominantly focused on quasi-static or fully actuated systems, thereby avoiding the complexities associated with underactuated and dynamically unstable manipulation scenarios.

\section{Methods}
\textbf{Task Formulation.}
In this paper, we investigate rope whipping using one-dimensional deformable objects as a representative example. The task requires hitting a specified goal location in the air with the tip of a rope that is attached to a two-joint robotic arm.
Let  $\bm{Q} = \{\boldsymbol{q}_1, \boldsymbol{q}_2, \dots, \boldsymbol{q}_N\}  \in \mathbb{R}^{N\times D}$  represent the $N$ squences of the action vector of the robotic system with $D$ DoFs, which in this context specifically refers to the robot's joint angles. Let $\mathcal{F}: \mathbb{R}^{N\times D} \to \mathbb{R}^3$ denote the forward dynamics mapping, which transforms a given action $\boldsymbol{q}$ into a Cartesian-space position $\mathbf{p} = (x, y, z)^\top \in \mathbb{R}^3$:
\begin{equation}
 \mathcal{H}(x, y, z) = \bm{Q}.
\end{equation}
The objective of goal-conditioned dynamic manipulation is to learn or approximate an inverse mapping $\mathcal{H}: \mathbb{R}^3 \to \ \mathbb{R}^{N\times D}$, such that:

\begin{equation}
\mathcal{H}(\mathbf{p}) = \bm{Q}, \quad \text{where } \mathcal{F}(\bm{Q}) \approx \mathbf{p}.
\end{equation}
Here, $\mathcal{F}$ denotes the forward dynamics mapping from the robot's action space $\bm{Q}$ to a position $\mathbf{p} \in \mathbb{R}^3$ in Cartesian space. The goal is to find the optimal action $\bm{Q}$that drives the end-effector or deformable object to reach a desired goal location $\mathbf{p}$ in 3D space.

\begin{figure*}[t]
\centering
\includegraphics[width=0.95\linewidth]{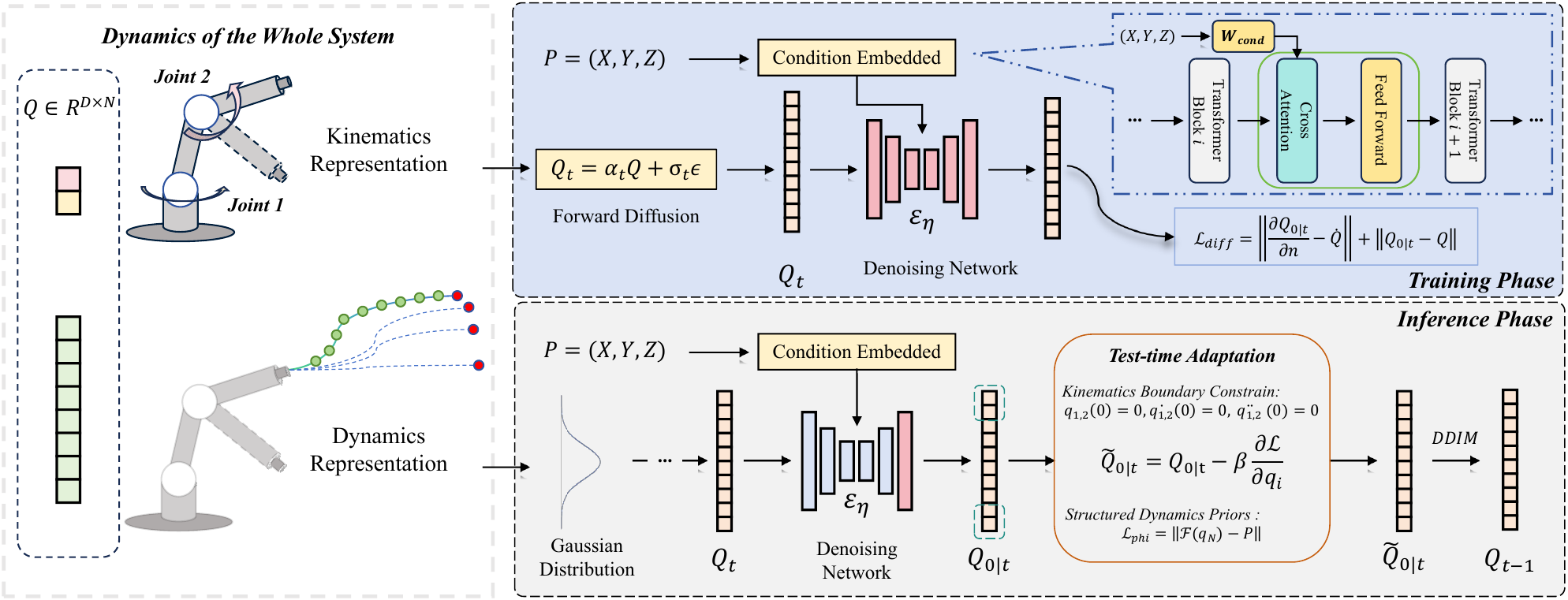} 
\vspace{-0.2cm}
\caption{Pipeline of the Proposed DIDP Framework.
$D$ denotes the DoF of the entire system, and $N$ the length of the action sequence. In the denoising network $\epsilon_\eta$, the red blocks represent learnable parameters, while blue blocks indicate frozen modules. During the training phase, all parameters are optimized jointly. During the inference phase, we employ a test-time adaptation strategy by fine-tuning only the final projection layer.}
\vspace{-0.5cm}
\label{fig:main}
\end{figure*}

\subsection{Inverse Dynamics Modeling}

\textbf{Reduced-Order Modeling.}
Most prior works rely on traditional FEM to model deformable dynamics and generate simulation and training data. In FEM, the deformable body is discretized into a dense mesh of nodes, where the position field in $\mathbb{R}^3$ is interpolated over these nodes, resulting in a high-dimensional system with a large number of DoFs. This high dimensionality significantly increases the difficulty of learning effective policies. Furthermore, previous works often model the grid and deformable components separately, which hinders the integration of physics-informed learning strategies~\cite{irp}. In this work, following~\cite{Anup2025reduced}, we adopt the GVS model to construct the simulator and prepare training data, achieving a significant reduction in the system’s state dimensionality. Below, we briefly revisit the GVS formulation.

The GVS model directly parameterizes the distributed strain field in  Lie algebra $\boldsymbol{\xi}_i \in \mathfrak{se}(3)$ using a finite set of generalized coordinates $\boldsymbol{q}_i \in \mathbb{R}^{D}$ and a set of basis functions $\Phi_{\xi_i}$. Based on this formulation, the configuration $\mathbf{g} \in SE(3)$ in the grid part can be recursively computed:
\begin{equation}
\boldsymbol{\xi}_i = \Phi_{\xi_i} \boldsymbol{q}_i + \boldsymbol{\xi}_i^*, \quad \mathbf{g}_i = \mathbf{g}_{i-1} \exp(\hat{\boldsymbol{\xi}_i})
\label{eq: grid}
\end{equation}
Here, $\boldsymbol{\xi}_i^*$ denotes a reference strain field, which can incorporate constraints such as inextensibility. The operator $\hat{\boldsymbol{\xi}}_i$ maps from $\mathfrak{se}(3)$ to $SE(3)$. This formulation enables the soft link configuration using only essential deformation modes. For the deformable objects parts, the $\mathbf{g} \in SE(3)$ can be recursively computed as following:
\begin{equation}
\mathbf{g}_i = \mathbf{g}_{i-1}\exp(\widehat{\Omega}_i),
\label{eq: soft}
\end{equation}
where $\widehat{\Omega}_i$ is computed via a truncated Magnus expansion of the strain field. Compared to FEM, which requires a large number of nodal displacements to capture fine-scale deformation, the strain-based GVS model achieves comparable accuracy with significantly fewer DoFs. Using $\boldsymbol{q}$ and the differential kinematic equations, the generalized Lagrangian equation of the robots dynamics is derived \cite{Anup2025reduced}:
\begin{equation}
\mathbf{M}(\boldsymbol{q}) \ddot{\boldsymbol{q}} + \mathbf{C}(\boldsymbol{q}, \dot{\boldsymbol{q}}) \dot{\boldsymbol{q}} + \mathbf{K}(\boldsymbol{q}) = \mathbf{B}(\boldsymbol{q}) \mathbf{u} + \mathbf{F}_{\text{ext}},
\end{equation}

\textbf{On the Choice of the Learning Space. } In many prior works on deformable object manipulation, the two joints of the robotic arm are treated as controllable, while the deformable object itself is considered uncontrollable. As a result, these methods often take observations of the deformable object as input and predict actions for the robotic joints. However, this decoupled formulation limits the expressiveness of the learned inverse dynamics, as the input-output mapping is not sampled from the full system's inverse dynamics. In this paper, we propose to jointly model the robot and the deformable object as a unified system. By directly predicting the joint angles $\bm{q}$, which include both the robotic arm and deformable components, our approach captures the full inverse dynamics more effectively and leads to improved learning performance. Thus, we consider the 20-DoF $\bm{Q} = \{\boldsymbol{q}_1, \boldsymbol{q}_2, \dots, \boldsymbol{q}_N\}  \in \mathbb{R}^{N\times 20}$ to describe the whole dynamics. Each vector $\boldsymbol{q}_i$ represents a 20-DoF action at $i-th$ squence of the overall actions. Where the $N$ denote the length of the sequence of the actions and $D$ denote the DoF of the whole dynamics systems, in which the first 2 DoF is used for the grid parts and the following are used to discribe the deformable objects parts.

\textbf{Diffusion-based Training.}
The diffusion model represents data through a stochastic process. Let $\bm{Q}_{0}$ denote a real robotic action, the forward diffusion process aims to generate a sequence of noisy latent variables $\bm{Q}_1,\bm{Q}_2, ..., \bm{Q}_T$ using a Markovian process, defined as:
\begin{equation}
    \begin{aligned}
        \bm{Q}_t = \alpha_t \bm{Q}_0 + \sigma_t\epsilon ,
    \end{aligned}	
    \label{eq: c-ddpm-forward}
\end{equation}
where $ \alpha_t \in(0,1)$ represents the noise schedule, $\sigma_t$ denotes the covariance at $t$, and $\epsilon$ represents the Gaussian noise. During training, the diffusion loss aims to maximize the likelihood of the sample results by using the variational lower bound. In the reverse process, we incorporate the position of the goal condition $ \bm{p}$ as the conditioning of the score functions $\epsilon_\eta(\bm{Q}_t, \bm{p}, t)$ and predict the robotic actions iteratively:

In the vanilla diffusion model\cite{dhariwal2021diffusion}, the denoising network is trained to predict the Gaussian noise and the recover the clean data using:
\begin{equation}
    \begin{aligned}
      \bm{Q}^\eta_{0|t}= \frac{\bm{Q}_t - \sigma_t\epsilon_\eta(\bm{Q}_t, \bm{p}, t)}{\alpha_t} .
    \end{aligned}	
    \label{eq: c-ddpm-reserved}
\end{equation}
Our diffusion policy is designed to directly predict the clean action sequence $\bm{Q}^\eta_{0|t}$. In addition, the angular velocity term $\bm{Q}_d$ is incorporated to enable second-order training, allowing the model to better capture dynamic behaviors:
\begin{equation}
    \begin{aligned}
    \mathcal{L}_{\text{diff}}&= \lambda_{Q}\|\bm{Q} -\bm{Q}^\eta_{0|t}\|_2^2 + \lambda_{Q_d}\|\bm{Q}_d -\dot{\bm{Q}}^\eta_{0|t}\|_2^2 ,\\
    \end{aligned}
\end{equation}

\begin{comment}
    \subsection{Network Designing}
We adopt a Transformer-based architecture as the denoiser of the diffusion policy, incorporating both cross attention and causal attention to effectively model the temporal structure and goal-conditioning in action prediction tasks.

\textbf{Cross Attention for Conditioning. } To ensure the model generates goal-directed actions, we employ \textit{cross attention} layers that inject goal information into each denoising step. 

\textbf{Autoregressive Action Prediction. } 
From a sequential modeling perspective, the generation process naturally follows an \textit{autoregressive factorization} of the conditional distribution:
\begin{equation}
    p(\bm{Q}_t \mid \bm{P}) = \prod_{i=1}^{T} p(\boldsymbol{q}_i \mid \boldsymbol{q}_{<i}, \bm{P})
\end{equation}

To capture this structure within our Transformer-based denoiser, we employ \textbf{causal self-attention}, which ensures that each token $\boldsymbol{q}_i$ attends only to itself and preceding tokens $\{\boldsymbol{q}_1, \dots, \boldsymbol{q}_{i-1}\}$. This structure is particularly suitable for modeling sequences of actions, where temporal causality must be strictly preserved for realistic and physically valid generation.

\end{comment}

\subsection{Differential Dynamics Prior from GVS model.} 
For goal-conditioned dynamic manipulation of deformable objects, it is essential to impose constraints in Cartesian space to ensure that the robot's joint movements correspond to the desired deformation behavior. Achieving this requires the system dynamics to be differentiable, enabling effective gradient-based policy optimization.

Revisiting the GVS model introduced in \cref{eq: grid} and \cref{eq: soft}, the position of any point on the end-effector can be directly obtained via $\bm{p} = \mathbf{g}_N(1\!:\!3,4)$. The dynamics prior aims to enforce consistency between the predicted pose $g^\eta_N$ and the goal pose $\tilde{g}_N$. To this end, we define the following loss function:
\begin{equation}
\mathcal{L}_{\text{pos}}= \left\| \log\left( \tilde{g}_N^{-1} \cdot g^\eta_N \right) \right\|^2,
\end{equation}
which penalizes discrepancies between the predicted and goal transformations in the Lie algebra. This loss is differentiable proved by the following corollary.

\begin{corollary} 
\label{cor:differentiable}
$\mathcal{L}_\text{pos}$ is differentiable with respect to the joint variables $\mathbf{q}_i$, i = 1, ..N.
\end{corollary}
\textbf{Proof.} 
Let:\[\boldsymbol{\delta} = \log\left( \tilde{\mathbf{g}}_N^{-1} \cdot \mathbf{g}^\eta_N \right)\]
Using the chain rule and the differential of the Lie logarithm, the gradient of the loss function with respect to joint variables $q_i$ can be approximated as:
\begin{equation}
\frac{\partial \mathcal{L}_{\text{pos}}}{\partial q_i} = 2 \boldsymbol{\delta}^\top \cdot \frac{\partial \boldsymbol{\delta}}{\partial q_i}  \approx 2 \boldsymbol{\delta}^\top \cdot \mathbf{J}_r^{-1}(\boldsymbol{\delta}) \cdot \frac{\partial \mathbf{g}_N}{\partial q_i}, \quad i = 1, ..N
\end{equation}
where $\mathbf{J}_r^{-1}(\cdot)$ denotes the inverse of the right Jacobian of the Lie group $\mathrm{SE}(3)$. Thus, the differentiability of $\frac{\partial \mathbf{g}_N}{\partial q_i}$ is sufficient for enabling gradient-based policy optimization. Recall from \cref{eq: grid} and \cref{eq: soft} that the dynamics of the grid and soft segments are modeled as:
\begin{equation}
\mathbf{g}_N = \exp(\hat{\boldsymbol{\xi}}_1) \cdot \exp(\hat{\boldsymbol{\xi}}_2) \cdot \exp(\widehat{\Omega}_3) \cdots \exp(\widehat{\Omega}_N).
\end{equation}
$\widehat{\Omega}_i$ can be computed via a truncated Magnus expansion. A fourth-order approximation is given by:
\begin{equation}
\widehat{\Omega}_i= \frac{H}{2}(\xi^1_i + \xi^2_i) + \frac{\sqrt{3} H^2}{12} [\xi^1_i, \xi^2_i],
\end{equation}
with $H$ being the discretization step size and $[\cdot,\cdot]$ denoting the Lie bracket capturing non-commutative effects. Here, $\xi^1_i$ and $\xi^2_i$ are  Zannah collocation points along the soft body. 
For the grid components with $i = 1, 2$, The partial derivative of the end-effector pose $\mathbf{g}_N$ with respect to the generalized coordinate $\boldsymbol{q}_1$ is given by:
\begin{equation}
\begin{aligned}
\frac{\partial \mathbf{g}_N}{\partial \boldsymbol{q}_1}
&= \left( \frac{\partial \exp(\hat{\boldsymbol{\xi}}_1)}{\partial \boldsymbol{q}_1} \right)
\cdot \exp(\hat{\boldsymbol{\xi}}_2)
\cdot \prod_{i=3}^{N} \exp(\widehat{\Omega}_i) \\
&= 
\exp(\hat{\boldsymbol{\xi}}_1), \cdot \mathbf{J}_l(\boldsymbol{\xi}_1) \cdot \Phi_{\xi_1} \cdot \exp(\hat{\boldsymbol{\xi}}_2) \cdot \prod_{i=3}^{N} \exp(\widehat{\Omega}_i)
\end{aligned}
\end{equation}
where $ \mathbf{J}_l(\boldsymbol{\xi}_1)$ is the left Jacobian associated $\boldsymbol{\xi}_1$.
For the deformable segments, indexed by $i = 3, 4, \ldots, N$, the partial derivative of the end-effector pose $\mathbf{g}_N$ with respect to the local generalized coordinate $\boldsymbol{q}_i$ is given by:
\begin{equation}
\frac{\partial \mathbf{g}_N}{\partial \boldsymbol{q}_i}
=
\left( \prod_{k=1}^{2} \exp(\widehat{\xi}_k) \right)\left( \prod_{k=3}^{i-1} \exp(\widehat{\Omega}_k) \right)
\cdot \mathbf{J}_{\text{l}}(\Omega_i)
\cdot \frac{\partial \Omega_i}{\partial \boldsymbol{q}_i}
\cdot \exp(\widehat{\Omega}_i)
\cdot \left( \prod_{k=i+1}^{N} \exp(\widehat{\Omega}_k) \right).
\end{equation}
$\mathbf{J}_{\text{l}}(\Omega_i)$ denotes the left Jacobian associated with $\Omega_i$. Assuming that the basis components $\boldsymbol{\xi}_i^1$ and $\boldsymbol{\xi}_i^2$ are differentiable with respect to $\boldsymbol{q}_i$, the partial derivative $\frac{\partial \Omega_i}{\partial \boldsymbol{q}_i}$ is well-defined and exists.

\subsection{The Physical Informed Test-Time Adaptation.}
In many real-world domains, especially in robotics and physical trajectory generation, diffusion models pre-trained via imitation learning can produce behaviorally plausible samples but may fail to satisfy strict physical or structural constraints at inference time. To mitigate this issue without disrupting the learned dynamics, we propose a \textit{physically informed test-time adaptation} (PITA) strategy that introduces lightweight structural regularization during the sampling process.

\textbf{Kinematic Boundary Condition.}
The zero-angle configuration typically corresponds to a standardized initial pose in many robotic platforms. Enforcing this condition ensures consistent and interpretable starting configurations across trajectories.
We need to enforce that $\bm{Q}(0)=0, \dot{\bm{Q}}(0)=0, \ddot{\bm{Q}}(0)=0$, which can be modeled as L2 regularization as a term of loss function $\mathcal{L}_{KBC}$.

\textbf{Test-Time Adaptation for Diffusion Sampling Process.}
Given a pre-trained diffusion model that predicts a denoised trajectory $\bm{Q}_{0|t}$ from a noisy sample $\bm{Q}_t$, we consider the incorporation of physical observations represented by $\bm{p}$. These observations depend on the clean trajectory through a known differentiable dynamics model $\bm{p} = \mathcal{F}(\bm{Q}_{0|t})$.

We aim to sample from the posterior distribution $p(\bm{Q}_t \mid \bm{p})$, which integrates the generative prior $p(\bm{Q}_t)$ with a physically grounded likelihood $p(\bm{p} \mid \bm{Q}_{0|t})$. For example, the likelihood can be instantiated as a soft constraint based on a differentiable loss function $\mathcal{L}(q_1, q_2)$, where $q_1, q_2$ are specific joint configurations extracted from $\bm{Q}_{0|t}$. This gives:
\[
p(\bm{p} \mid \bm{Q}_{0|t}) \propto \exp\left(-\mathcal{L}(q_1, q_2)\right).
\]

To guide the sampling process under this posterior, we apply the identity from score-based generative modeling to approximate the posterior score:
\begin{equation}
\begin{aligned}
    \nabla_{\bm{Q}_t} \log p(\bm{Q}_t \mid \bm{p}) &\approx \nabla_{\bm{Q}_t} \log p(\bm{Q}_t) + \nabla_{\bm{Q}_{t}} \log p(\bm{p} \mid \bm{Q}_{t})\\
   & =\nabla_{\bm{Q}_t} \log p(\bm{Q}_t) +  \frac{\partial \mathcal{L}(q_1, q_2)}{\bm{Q}_{0|t}} \cdot \frac{\partial \bm{Q}_{0|t}}{\partial \bm{Q}_t},
\end{aligned}
\label{eq:posterior_score}
\end{equation}
Since $\bm{Q}_{0|t}$ is the output of the diffusion policy, and the differentiability of $\frac{\partial \mathcal{L}(q_1, q_2)}{\partial \bm{Q}_{0|t}}$ has been established in  Corollary~\ref{cor:differentiable}, the full gradient $\frac{\partial \mathcal{L}(q_1, q_2)}{\partial \bm{Q}_t}$ can be computed via the chain rule by propagating through the denoiser network. To ensure stable test-time adaptation, we restrict gradient updates to only the final projection layer of the denoiser network. The final loss function for test-time adaptation $\mathcal{L}(q_1, q_2)$ is as following:
\begin{equation}
\mathcal{L}= \mathcal{L}_{\text{pos}} + \mathcal{L}_{\text{KBC}},
\end{equation}

Equation~\ref{eq:posterior_score} enables physically informed guidance at each sampling step by modifying either the sample $\bm{Q}_t$ or the model parameters, thereby enforcing consistency with domain-specific physical constraints during test-time inference.

\section{Experiments}
\subsection{Dataset}

\begin{figure*}[t]
\centering
\includegraphics[width=0.95\linewidth]{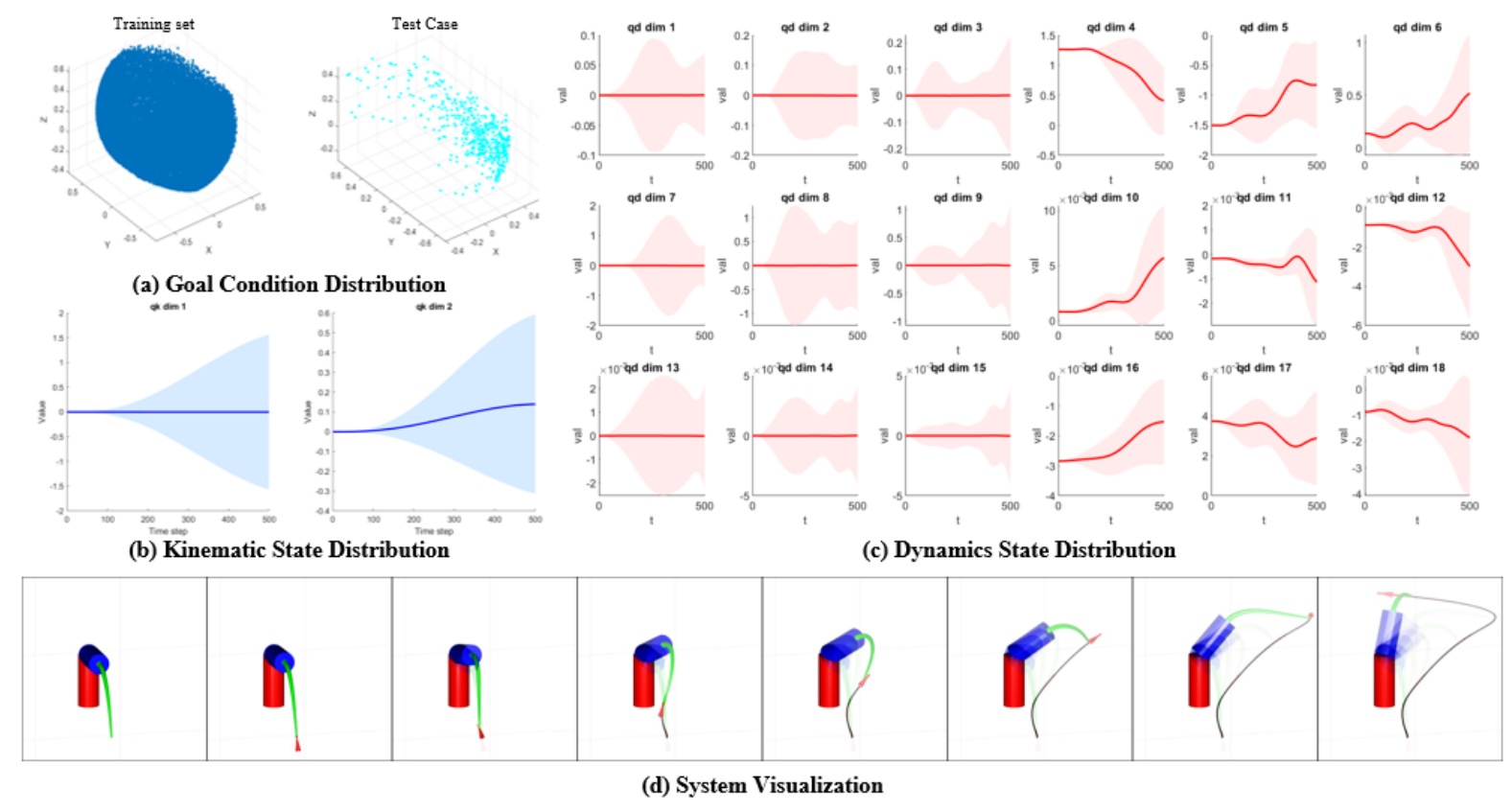} 
\vspace{-0.3cm}
\caption{Dataset overview. (a) shows the distribution of goal conditions in 3D space for both training and testing cases, demonstrating comprehensive spatial coverage. (b) and (c) depict the distributions of kinematic and dynamic states, respectively. The curves indicate the mean and standard deviation across time steps, based on a reduced representation with 20 DoFs to describe the entire system. (d) provides a visualization of the complete system within the simulation environment.}
\vspace{-0.5cm}
\label{fig:dataset}
\end{figure*}
\textbf{Overview.} 
Our benchmark dataset comprises $N=55{,}000$ high-fidelity simulated trajectories of a whip-like continuum soft robot, as shown in Figure~\ref{fig:dataset}. Each trajectory is generated by first sampling an actuation input matrix $\theta \in \mathbb{R}^{2\times4}$ uniformly from the operational input space, then simulating the system's dynamic response over $t\in[0,T]$ using a Cosserat rod-based physics engine. Simulations run for $T=0.5$s using a fixed timestep $\Delta t = 0.001$s (RK4), resulting in $L=500$ time steps. At each time step, we record the complete configuration state $\boldsymbol{Q}\in \mathbb{R}^{N\times D}$, velocity state $\dot{\boldsymbol{Q}}\in \mathbb{R}^{N\times D}$, and the Cartesian positions and velocities of 21 evenly spaced material points. % This yields time-series data tensors of shape $L\times21\times3$ for both positions and velocities.

\textbf{Sampling Strategy.} 
To comprehensively capture the spatial and temporal diversity of soft continuum robot dynamics, we propose a structured yet randomized sampling strategy that generates large-scale sequences. Each control input is represented as a parameter matrix $\boldsymbol{q} \in \mathbb{R}^{2 \times 4}$, where the two rows correspond to the independently actuated joints of the robot, and the four columns encode piecewise-constant actuation commands over four equal-duration temporal segments.

Control inputs are structured as $\boldsymbol{q} \in \mathbb{R}^{2 \times 4}$, with each row encoding actuation for one joint over four temporal segments. Candidate matrices are drawn from joint-specific uniform distributions: $\boldsymbol{q}_1 \sim \mathcal{U}[-\pi, \pi]$ and $\boldsymbol{q}_2 \sim \mathcal{U}[-\pi/2, \pi/4]$. To ensure stability and physical validity, trajectories with numerical errors or divergence are filtered. Valid samples are simulated using a stiff-aware solver (\texttt{ODE15s}) with Jacobian-based evaluations.

\textbf{Evaluation Metrics.} We use the Euclidean distance between the rope's tip and the goal position in Cartesian space as the primary evaluation metric. To assess performance under varying precision requirements, we define success based on distance thresholds of 10\,cm, 5\,cm, 2\,cm, and 1\,cm. The success rate is computed as the percentage of trials in which the rope tip falls within each specified threshold. This multi-level evaluation provides a comprehensive assessment of both coarse and fine-grained control capabilities.

\subsection{Validation on the Learning Space.}
\textbf{Experimental Settings.} We evaluate the effectiveness of different learning spaces using a diffusion policy with a Transformer-based denoiser. We compare two settings:

\textit{Kinematics}: The policy is trained solely on the kinematic states, i.e., it learns to replicate the behavior of the robot’s actuated joints without considering the deformable rope.

\textit{Kinematics + Dynamics}: The policy is trained on both the robot and the rope using the reduced-order dynamics model, thereby capturing the full system behavior.

\textbf{Experimental Analyisis.} Note that the kinematics-only model lacks dynamic state representations, and therefore cannot leverage our test-time adaptation strategy. Despite this, as shown in \cref{tab: Learning_Space}, extending the learning space to include dynamics yields substantial performance improvements. This highlights the value of incorporating physically meaningful inverse dynamics: it not only enhances interpretability but also significantly boosts imitation learning performance.
\begin{table}[t]
\centering
\vspace{-0.4cm}
\resizebox{0.9\textwidth}{!}{\begin{tabular}{cc|cc|c|cccc}
\toprule
\multirow{2}{*}{Kinematic} & \multirow{2}{*}{Dynamics}        & \multirow{2}{*}{DDIM}       & \multirow{2}{*}{TTA}       & \multirow{2}{*}{Distance} & \multicolumn{4}{c}{Success Rate}   \\ \cline{6-9}
&         &         &  &            & 10 cm  & 5cm    & 2 cm   & 1 cm  \\ \midrule
\checkmark    &                           & \checkmark   &  & 0.065                        & 0.773  & 0.699  & 0.434  & 0.107          \\ 
\checkmark & \checkmark & \checkmark &                           & 0.041                        & 0.884  & 0.800  & 0.616  & 0.190 \\
\checkmark & \checkmark &                           & \checkmark & 0.036                     & 0.939  & 0.843  & 0.623  & 0.208 \\ \midrule
\end{tabular}}
\caption{Quantitative comparison of learning spaces. The extending the learning space to include dynamics leads to significant improvements in task performance.}
\vspace{-0.8cm}
\label{tab: Learning_Space}
\end{table}

\subsection{Validation on the Learning Strategy.}
\textbf{Experimental Settings.} We evaluate the effectiveness of different pretraining loss formulations by comparing three learning strategies: Imitation Learning (IL), Trajectory Optimization (TO), and their combination (IL+TO).

\textit{IL} denotes standard Imitation Learning, where the diffusion policy is trained via behavior cloning to match the demonstrated full-system dynamics. The training objective includes the loss term $\mathcal{L}_{\text{diff}}$ to directly fit expert trajectories.

\textit{TO} denotes Trajectory Optimization-based learning, where the diffusion policy is optimized to produce action sequences that drive the rope’s end-effector toward the goal positions. This strategy emphasizes goal-directed behavior rather than mimicking demonstrations.

\textit{IL+TO} represents our proposed hybrid strategy. It first pretrains the diffusion policy using Imitation Learning for stable behavior generation, and then fine-tunes it with Trajectory Optimization to improve goal-reaching accuracy.

\textbf{Experimental Analysis.}  
Quantitative comparisons are conducted across several goal-reaching scenarios are shown in Table~\ref{tab: IL_TO}. Results show that: \textbf{IL} provides stable behavior generation but lacks task-specific precision, often failing to reach the desired goals.
\textbf{TO} improves success in specific cases but is unstable and inefficient when trained from scratch due to the high dimensionality and dynamics complexity. \textbf{IL+TO} achieves the best performance by combining the stability of imitation with the adaptability of optimization, showing significantly higher success rates and better policy generalization across unseen goals. 
We also visualize the actions of the two robotic joints. As shown in Figure~\ref{fig:IL_TO}, using \textbf{TO} independently disrupts the behavior cloning of the system dynamics, leading to degraded motion consistency.
These findings suggest that while imitation learning provides a good initialization, fine-tuning with goal-directed optimization is essential for dynamic manipulation tasks involving deformable objects.
 
\begin{table}[ht]
\centering
\vspace{-0.35cm}
\begin{minipage}{0.49\textwidth}
\centering
\resizebox{\textwidth}{!}{
\begin{tabular}{cc|c|cccc}
\toprule
\multirow{2}{*}{IL} & \multirow{2}{*}{TO}  & \multirow{2}{*}{Distance} & \multicolumn{4}{c}{Success Rate}   \\ \cline{4-7}
    &   &            & 10 cm  & 5 cm   & 2 cm   & 1 cm  \\ \midrule
\checkmark &                & 0.041                        & 0.884  & 0.800  & 0.616  & 0.190  \\
          & \checkmark                     & 0.206   & 0.155  & 0.068  & 0.018  & 0.007 \\
\checkmark & \checkmark                    & 0.036                     & 0.939  & 0.843  & 0.623  & 0.208 \\ \midrule
\end{tabular}
}
\caption{Quantitative comparison of learning strategies. TO improves the success rate of IL in dynamic manipulation of deformable objects but fails to learn an efficient policy independently.}
\label{tab: IL_TO}
\end{minipage}
\hfill
\begin{minipage}{0.49\textwidth}
\centering
\centering
\includegraphics[width=0.95\linewidth]{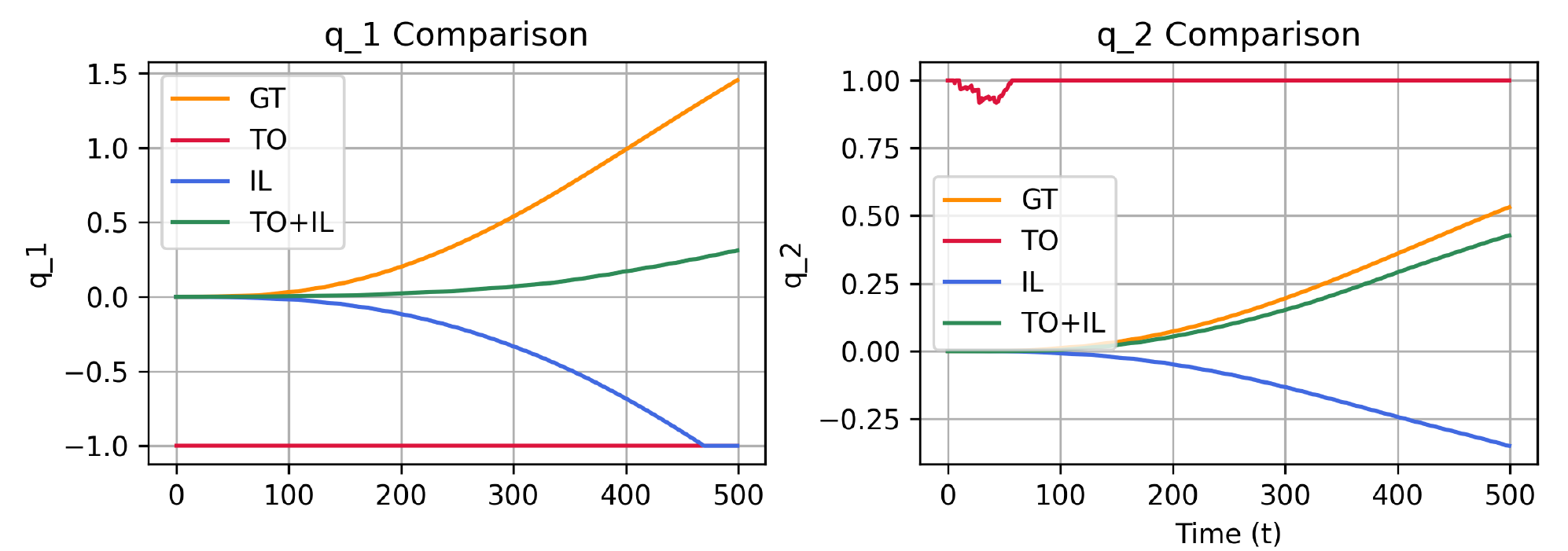}
\captionof{figure}{A case study comparing learning strategies. TO improves IL's action accuracy but fails to produce correct actions on its own.}
\label{fig:IL_TO}
\end{minipage}
\vspace{-0.5cm}
\end{table}

\subsection{Ablation study on the Test-time Adaption.}

\begin{figure*}[t]
\centering
\includegraphics[width=0.95\linewidth]{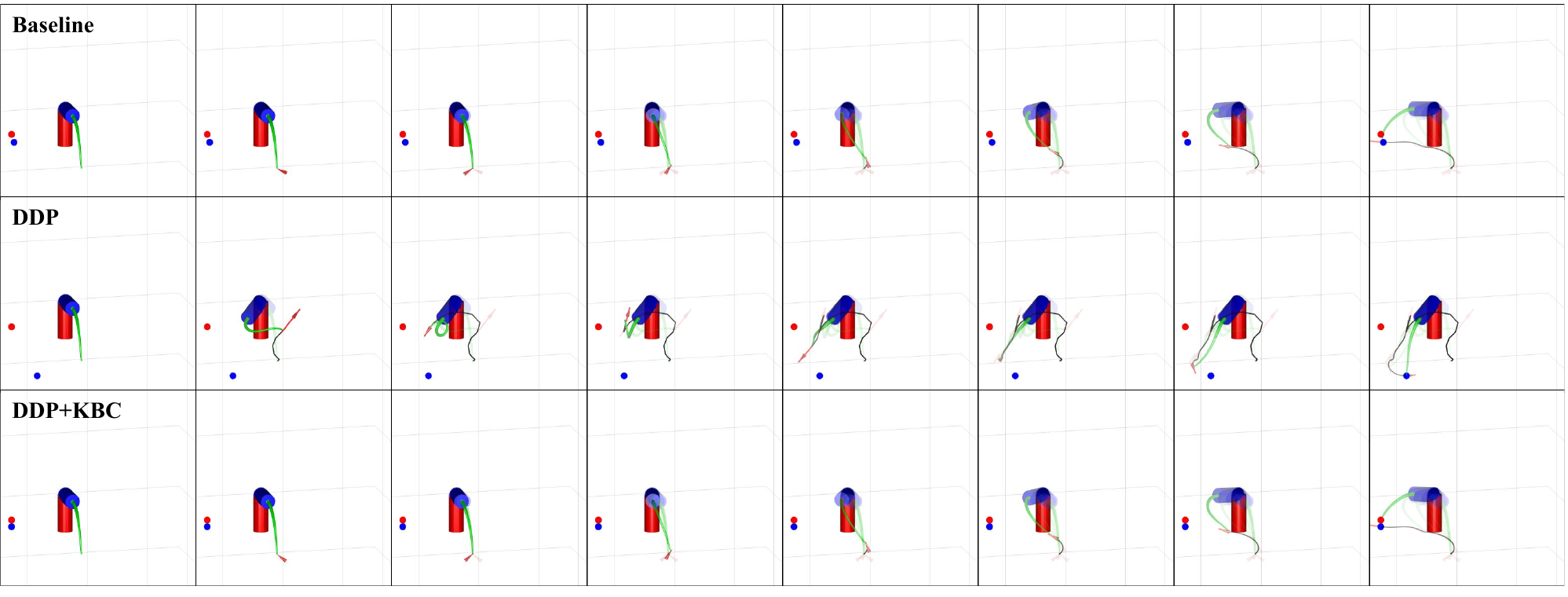} 
\caption{Visualization of the manipulation process. The baseline represents the diffusion policy without test-time adaptation. \textbf{DDP} denotes test-time adaptation using only the differentiable dynamics prior, while \textbf{KBC} indicates adaptation incorporating the kinematic boundary condition.}
\vspace{-0.5cm}
\label{fig:tta_ablation}
\end{figure*}
\textbf{Experimental Settings.} We evaluate the effectiveness of the proposed test-time adaptation strategy by comparing different configurations:

\textit{Tuning Strategy.} We examine two approaches for updating model parameters during test-time adaptation. \textbf{Full Finetune} indicates that all parameters are fine-tuned during adaptation, while \textbf{Project Finetune} refers to freezing most of the model and updating only the final projection layer.

\textit{Physical Priors.} We assess the influence of incorporating physical constraints during test-time adaptation. \textit{DDP} denotes the use of a Differential Dynamics Prior to guide the adaptation process, and \textit{KBC} refers to enforcing Kinematic Boundary Conditions to ensure physically consistent outputs.

\textbf{Experimental Analysis.}
We observe that incorporating KBC is essential for maintaining consistency at the trajectory's initial state. As shown in Table~\ref{tab: phi_prior}, without KBC, test-time adaptation using DDP becomes unstable, often producing physically implausible motions and reduced task performance.
Moreover, different adaptation strategies during fine-tuning yield varied results. As shown in Table~\ref{tab: Tuning}, full finetuning incurs nearly 40\% additional inference time per sample while leading to a drop in overall performance. This degradation arises because updating the entire network often disrupts the behaviors learned during the imitation phase, negatively affecting both convergence and generalization. In contrast, restricting adaptation to the final projection layer offers a more favorable trade-off between flexibility and stability.

\begin{table}[ht]
\centering
\vspace{-0.4cm}
\begin{minipage}{0.49\textwidth}
\centering
\resizebox{\textwidth}{!}{
\begin{tabular}{cc|c|cccc}
\toprule
\multirow{2}{*}{DDP} & \multirow{2}{*}{KBC}             & \multirow{2}{*}{Distance} & \multicolumn{4}{c}{Success Rate}   \\ \cline{4-7}
    &   &             & 10 cm  & 5cm    & 2 cm   & 1 cm  \\ \midrule
      &                                     & 0.057                        & 0.884  & 0.800  & 0.616  & 0.190    \\ 
\checkmark &                     & 0.189                        & 0.304  & 0.124  & 0.038  & 0.020 \\
\checkmark &\checkmark     & 0.036                     & 0.939  & 0.843  & 0.623  & 0.208 \\ \midrule
\end{tabular} 
}
\caption{Ablation study on the choice of physical priors. KBC plays a crucial role in constraining the initial boundary states. When omitted during test-time adaptation with DDP, the policy suffers from unstable predictions, leading to a significantly lower success rate.}
\label{tab: phi_prior}

\end{minipage}
\hfill
\begin{minipage}{0.49\textwidth}
\centering
\resizebox{\textwidth}{!}{\begin{tabular}{c|c|c|cccc}
\toprule
\multirow{2}{*}{Tuning Strategy}  & \multirow{2}{*}{Time (s)}  & \multirow{2}{*}{Distance} & \multicolumn{4}{c}{Success Rate}   \\ \cline{4-7}
    &         &  & 10 cm  & 5 cm   & 2 cm   & 1 cm  \\ \midrule
Full Finetune  &  14.22  &  0.096     &  0.864 &0.678   &0.146   &0.020  \\
Project Finetune  & 10.39  & 0.036                     & 0.939  & 0.843  & 0.623  & 0.208 \\ \midrule
\end{tabular}}
\caption{Ablation study on the tuning strategy. Full finetuning tends to overwrite the knowledge learned during imitation pretraining, resulting in instability in the final policy. In contrast, projection-layer finetuning, which updates fewer parameters, preserves the pretrained behavior while enabling effective adaptation.}
\label{tab: Tuning}
\end{minipage}
\vspace{-0.5cm}
\end{table}

\subsection{Limitations.}
This work focuses primarily on learning the inverse dynamics of the entire system under a specific deformable object setup. However, it does not account for variations in object shape, material properties, or other physical characteristics that may significantly impact system behavior. As a result, the learned policy is tailored to a single type of deformable object, limiting its generalizability. In future work, we aim to address this limitation by expanding the dataset to include diverse deformable objects and incorporating object-specific attributes into the learning process. % Additionally, we plan to explore multi-task policy learning frameworks that can generalize across different object configurations, paving the way toward a more unified and robust manipulation strategy.

\section{Conclusion}

In this work, we present DIDP, a Dynamics-Informed Diffusion Policy framework for goal-conditioned dynamic manipulation of deformable objects in 3D environments. Our approach combines a reduced-order modeling formulation with a diffusion-based policy that incorporates physical priors through test-time adaptation. By modeling the inverse dynamics within a compact, task-relevant latent space, DIDP achieves efficient and physically grounded action generation. Extensive experiments demonstrate that DIDP outperforms existing baselines in terms of both accuracy and robustness. We believe this work offers a promising direction for bridging efficient learning and physically control in dynamic robotic manipulation.

%%%%%%%%%%%%%%%%%%%%%%%%%%%%%%%%%%%%%%%%%%%%%%%%%%%%%%%%%%%%

\setlength{\bibsep}{5pt}
%\bibliographystyle{plainnat}
%{\small \bibliography{main.bib}}
%{
{\small
\bibliographystyle{plain}
\bibliography{main}
}

\end{document}